\documentclass[letterpaper]{article} 
\usepackage[table,xcdraw]{xcolor}
\usepackage{color}
\usepackage{aaai25}  
\usepackage{times}  
\usepackage{helvet}  
\usepackage{courier}  
\usepackage{amsmath}
\usepackage[hyphens]{url}  
\usepackage{graphicx} 
\usepackage{natbib}  
\usepackage{caption} 
\usepackage{algorithm}
\usepackage{algorithmic}

\usepackage{nameref}
\usepackage[inline]{enumitem}
\usepackage[many]{tcolorbox}

\usepackage{newfloat}
\usepackage{listings}
\DeclareCaptionStyle{ruled}{labelfont=normalfont,labelsep=colon,strut=off} 
\floatstyle{ruled}
\newfloat{listing}{tb}{lst}{}
\floatname{listing}{Listing}
\usepackage{booktabs}

\newcommand{\jf}[1]{\textcolor{red}{[\textbf{JF:} #1]}}
\newcommand{\ourapproach}{\texttt{LLM-ACTR}}

\newcommand{\sref}[1]{\textit{\nameref{#1}}}

\definecolor{codegreen}{rgb}{0,0.6,0}
\definecolor{codegray}{rgb}{0.5,0.5,0.5}
\definecolor{codepurple}{rgb}{0.58,0,0.82}
\definecolor{backcolour}{rgb}{0.95,0.95,0.92}

\title{Cognitive LLMs: Towards Integrating Cognitive Architectures and Large Language Models for Manufacturing Decision-making}

\author{
    Siyu Wu\textsuperscript{\rm 1}\ \quad
    Alessandro Oltramari\textsuperscript{\rm 2,3}\ \quad
    Jonathan Francis\textsuperscript{\rm 2,3}\ \quad
    C. Lee Giles\textsuperscript{\rm 1}\ \quad
    Frank E. Ritter\textsuperscript{\rm 1}\\
}
\affiliations{
    \textsuperscript{\rm 1}College of Information Sciences and Technology, The Pennsylvania State University; University Park, USA\\
    \textsuperscript{\rm 2}Bosch Center for Artificial Intelligence; Pittsburgh, USA\\
    \textsuperscript{\rm 3}Carnegie Mellon University; Pittsburgh, USA\\
    {\tt \footnotesize \{sfw5621, clg20, frank.ritter\}@psu.edu, \{alessandro.oltramari, jon.francis\}@us.bosch.com}
}

\newtcolorbox{llmprompt}[3][]{%
    enhanced jigsaw, 
    breakable,      
    left=1cm,       
    right=1cm,      
    colframe    = #2!25,
    colback     = #2!10,
    coltitle    = #2!20!black,  
    title       = {#3},
    parbox=false,
}

\newtcolorbox{llmresponse}[3][]{%
    enhanced jigsaw, 
    breakable,      
    left=1cm,       
    right=1cm,      
    colframe    = #2!25,
    colback     = #2!10,
    coltitle    = #2!20!black,  
    title       = {#3},
    parbox=false,
}

\begin{document}

\maketitle

\begin{abstract}
Resolving the dichotomy between the human-like yet constrained reasoning processes of Cognitive Architectures and the broad but often noisy inference behavior of Large Language Models (LLMs) remains a challenging but exciting pursuit, for enabling reliable machine reasoning capabilities in production systems. Because Cognitive Architectures are famously developed for the purpose of modeling the internal mechanisms of human cognitive decision-making at a computational level, new investigations consider the goal of informing LLMs with the knowledge necessary for replicating such processes, e.g., guided perception, memory, goal-setting, and action. Previous approaches that use LLMs for grounded decision-making struggle with complex reasoning tasks that require slower, deliberate cognition over fast and intuitive inference---reporting issues related to the lack of sufficient grounding, as in hallucination. To resolve these challenges, we introduce \ourapproach, a novel neuro-symbolic architecture that provides human-aligned and versatile decision-making by integrating the ACT-R Cognitive Architecture with LLMs. Our framework extracts and embeds knowledge of ACT-R's internal decision-making process as latent neural representations, injects this information into trainable LLM adapter layers, and fine-tunes the LLMs for downstream prediction. Our experiments on novel Design for Manufacturing tasks show both improved task performance as well as improved grounded decision-making capability of our approach, compared to LLM-only baselines that leverage chain-of-thought reasoning strategies. 




\end{abstract}

\section{Introduction}
Large-capacity neural foundation models, such as Large Language Models (LLMs), have gained considerable popularity for a wide range of prediction and decision-making tasks, spanning applications, such as robotics and control, neural question-answering, scene understanding, code generation, mathematical reasoning, etc. LLMs are trained on massive datasets, can be used both as discriminative scoring functions as well as generative models, and their capacity allows them to accumulate and retain vast amounts of knowledge \cite{Brown2020LanguageModels, Andreas2022LanguageModels, dong2022survey, francis2022core, hu2023toward}. 
On the surface, typical usage of LLMs mirrors `\textit{system-1 reasoning processes}' \cite{Sloman1996, Hagendorff2023}, from the dual-process theory of human cognition \cite{wason1974dual, kahneman2011thinking}, which provide fast, intuitive, and automatic reasoning---underpinning tasks like navigating daily environments and making quick decisions. Advancements in multi-agent LLM frameworks as well as emergent capabilities such as in-context learning \cite{CodaFornoEtAl2024, dong2022survey} have enabled LLMs to employ more sophisticated reasoning strategies, such as `chain-of-thought' reasoning (CoT) \cite{Bhattamishra2023Understanding}. These capabilities facilitate LLMs' pursuit of `\textit{system-2 processes}' \cite{TverskyKahneman1974}, which involve slower, deliberate cognition and critical thinking for complex tasks \cite{Brown2020LanguageModels, WebbHolyoakLu2022}---essential for decision-making in realistic settings. While LLMs have shown promise in this area, key concerns remain, e.g., over discrepancies between LLM inference behavior and human reasoning \cite{BinzSchulz2023,Liu2024Large}, in analyses showing that LLMs prioritize fast and intuitive ``system-1" thinking over slower and deliberate analysis \cite{Hagendorff2023}, and over issues of insufficient grounding such as hallucination \cite{chakraborty2024hallucination}. These issues raise potential concerns about deployment settings where LLMs are left to perform inference, without having been first grounded on reliable knowledge sources or decision processes \cite{yao2022react}. 

To alleviate these issues, we propose \ourapproach, which shows improved decision-making capabilities over LLMs by integrating intermediate representations extracted from a well-establish neuro-symbolic system: the ACT-R cognitive architecture \cite{anderson2009can, ACTR}. ACT-R 
offers an integrated theory of the mind 
--- encompassing perception, memory, goal-setting, and action --- and has been pivotal in developing synthetic agents for learning and training \cite{anderson2019}. The representation extracted from ACT-R cognitive models serves as domain knowledge, infusing LLMs with decision-making augmentation. 
\begin{figure*}[ht]
\centering
\includegraphics[width=0.9\textwidth]{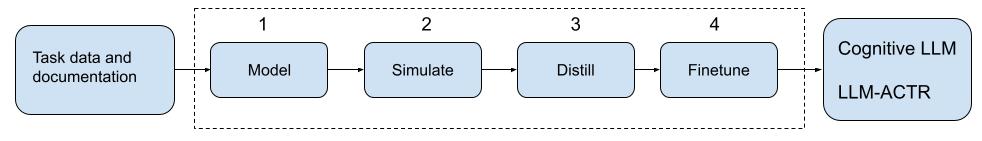}
\caption{Decision augmentation using a neural-symbolic cognitive architecture approach. (1) Tasks are modeled with cognitive architecture. (2) Cognitive model used to run stochastic simulation of task at scale. (3) Synthetic data are distilled from simulation and combined with prompt requests. (4) A fine tuning pipeline is used to calibrate open source LLM to perform decision augmentation for task in exercise.
} 
\label{fig:taskprocess}
\end{figure*}
\ourapproach~uses ACT-R models to represent human repeated decision-making with learning. We infuse ACT-R model's intermediate representations with the last hidden layers of open source LLM, and add a top classification layer for fine-tuning. The architecture is then deployed in unseen decision-making tasks. The LlaMa model family \cite{Touvron2023LLaMA} was selected for this research, due to its full accessibility to network architecture, including its pre-trained weights, and its proven efficacy in prior applications involving the extraction of the last hidden layer for predicting behavior discrepancies \cite{BinzSchulz2023a}. This approach integrates the ACT-R model's representation of human-like decision-making patterns into the LLM, enhancing its ability to make decisions that are both human-aligned and explainable. The fine-tuned LLM transcends mere prediction of human decisions for unseen problems. Significantly, it outlines a road-map for enabling high-level machine reasoning through cognitive neuro-symbolic systems \cite{Oltramari2023}. \ourapproach~leverages the strengths of both LLMs and CAs by using LLMs' natural language processing and generative capabilities, complemented by the human-aligned reasoning and explainability offered by CAs.

This paper presents a deployment case of \ourapproach~in manufacturing decision-making, demonstrating how this approach addresses the typically noisy inference behavior associated with off-the-shelf LLMs in real-world settings. The task is associated with the key aspect of DFM: enhancing product development and optimizing production system performance by improving time efficiency and reducing headcount costs \cite{Ulrich1993}. In this work, we introduce Cognitive LLMs, describe our proposed framework \ourapproach, and provide  experimental comparisons on a novel task with strong baselines.





\section{Related Work}
\label{sec:related_work}

\paragraph{Cognitive Decision Making}
The exploration of human cognition and decision-making processes has been a longstanding pursuit in cognitive science and artificial intelligence research \cite{gluck2006modeling}. Two prominent frameworks for cognitive modeling are ACT-R \cite{anderson2009can, Bothell2017} and Soar \cite{laird2019soar}: these frameworks serve as robust tools for simulating human behavior across various cognitive tasks. They are referred to as Cognitive architectures (CAs) \cite{laird2019soar, anderson1998}, reflecting a set of intertwined mechanisms to model human behavior 
and aiming for a unified representation of mind \cite{Newell1994}. CAs use task-specific knowledge to generate behavior. They represent various types of knowledge, including declarative (factual), procedural (how-to), and even recent advancements in perception and motor skills. This knowledge allows CAs to not only simulate behavior but also to explain it, both through direct examination and by tracing the reasoning steps involved in real-time (concurrent protocol). ACT-R encompasses perception, memory, goal-setting, and action, and it has been widely applied in decision-making tasks across psychology and computer science e.g.,\cite{BestLebiere2003, wu2024llms, Gonzalez2003}. 
ACT-R was chosen for this study to provide the intermediate representations of real time reasoning steps. However, the ACT-R model does not accept natural language as input and cannot generalize across different tasks, even within the same domain, which limits its flexibility for decision-making. In contrast, \ourapproach~combines the strengths of both LLMs and ACT-R models by leveraging the natural language processing and generative capabilities of LLMs, and making decisions that are grounded by those of ACT-R models.

\paragraph{Integration of Cognitive Architectures and LLMs}
CAs face limitations due to domain restrictions, which have hampered their broader application. They are incapable of processing natural language, they are limited to areas that can be described by logical rules, and they require a significant number of pre-defined rules to function. Intriguingly, LLMs \cite{Brown2020LanguageModels} offer potential solutions to mitigating these weaknesses. LLMs can process a variety of text inputs and are more flexible than rule-based systems. Additionally, they seem to learn rules implicitly, through pre-training, eliminating the need for manual rule creation. Hence, the notion of integrating CAs and LLMs is attractive, for leveraging the strengths of both approaches and thereby creating a more robust unified theory of computational models. This integration can take several forms, however, e.g., leveraging the implicit world knowledge of LLMs to replace the CAs' declarative knowledge mechanisms or to enhance their traditional symbolic mechanisms for procedural knowledge \cite{wray2024eliciting, kirk2024improving}. 
Additional research explores how principles from cognitive architectures can guide the design of LLM-based agent frameworks \cite{sumers2023cognitive}, demonstrating a comprehensive integration effort that spans from knowledge representation to interaction with the environment. However, to our knowledge, unlike these previous efforts that incorporate LLMs into CAs, there is currently no research focusing on assimilating the advantages of CAs into LLMs. In this paper, we leverage a cognitive architecture to ground the reasoning process and outputs of LLMs; by assimilating a neural representation of ACT-R model within LLMs, we aim to enhance LLMs' human alignment and explainability. 

\begin{figure}[htbp]
    \centering
    \includegraphics[width=\columnwidth]
    {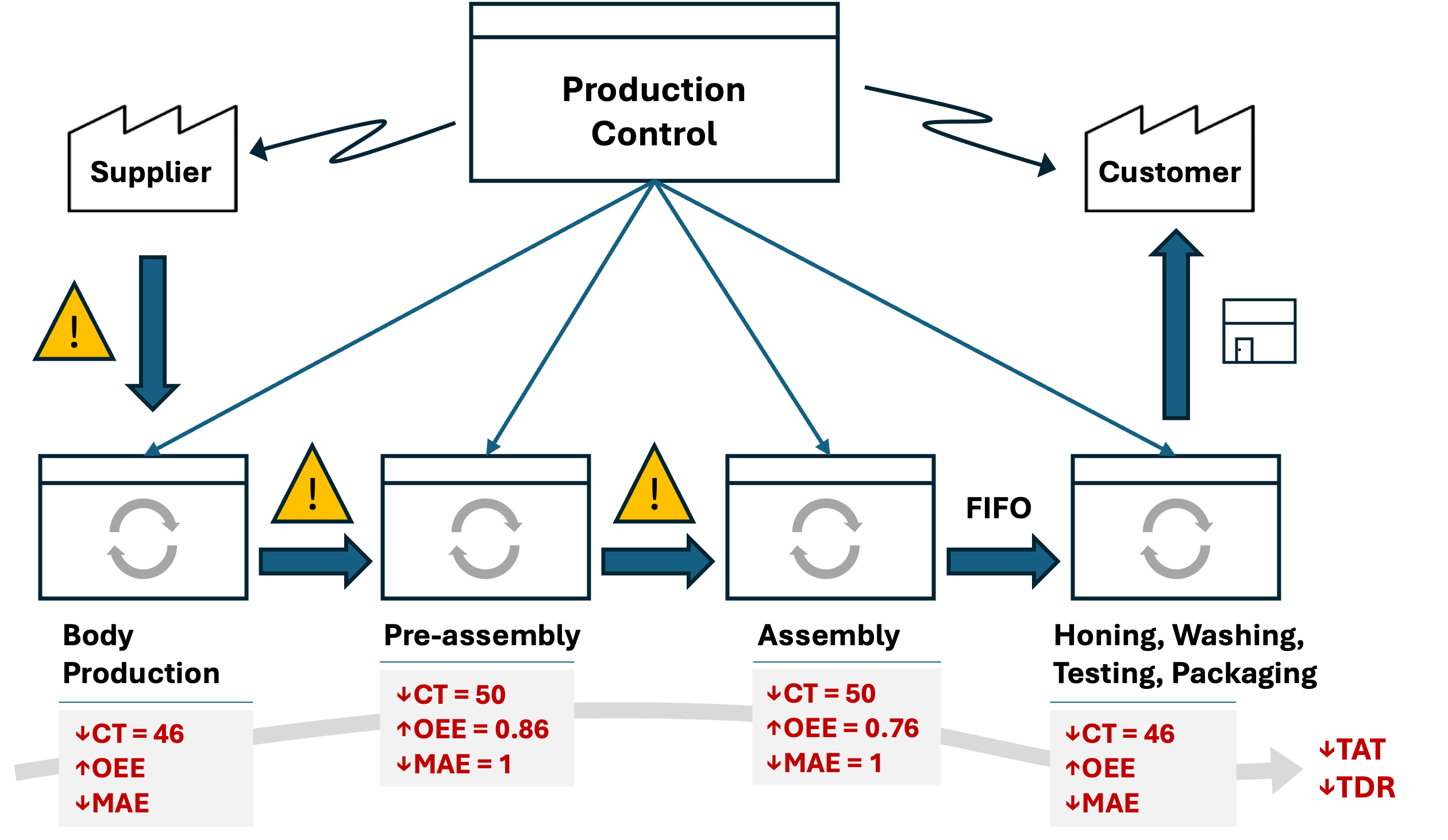}
    \caption{A \textit{Value Stream Map} of manufacturing process.}
    \label{fig:value_stream_map}
\end{figure}

\section{Problem Definition: Design for Manufacturing}
We define the terminology that constitutes our problem. The problem setting is a prototypical manufacturing production-line workflow, from supplier to customer, for which there exists a Value Stream Map (VSM; see Figure \ref{fig:value_stream_map}), which allows for tracking the efficiency at different sectors of the process and abstracts the overall problem for mathematical modeling and optimization. Key sectors include: Body Production, Pre-Assembly, Assembly, Honing, Washing, Testing, and Packaging. Early sectors pose potential efficiency problems in the workflow and may warrant optimization (triangles), while later stages are governed by \textit{First-In-First-Out} (FIFO) processes. 
The metrics at each stage include Cycle Time (CT), Overall Equipment Effectiveness (OEE), and Mean Absolute Error (MAE); the flow progresses through each stage, aiming for efficient operation, performance monitoring, and error minimization to ensure high-quality production output and timely customer delivery.

Focused on maintaining stable output for manufacturing plants, we consider plant managers' feedback alongside the VSM structure to define two decision-making problems that aim to reduce Total Assembly Time (TAT) while minimizing Total Defect Rate (TDR). An agent $\mathcal{G}$ is a predictive model that takes a natural language question $\mathcal{Q}$ as a prompt, along with $N$ snapshots of the sector-wise production flow data \{CT, OEE, MAE\}. In a \textit{single-facet decision-making problem}, $\mathcal{G}$ outputs a binary decision ($0$ or $1$) on which of two sectors, pre-assembly or assembly, requires a time reduction. In a more-challenging \textit{multi-faceted decision-making problem}, $\mathcal{G}$ should output the same binary decision as before, about which sector should be the optimization target, along with an optimization \textit{strategy} $S$. Here, $S$ is a strategy defined by one of several decision-making personas that govern manufacturing process management, which we refer to in the manuscript as `novice', `intermediate', and `expert'.


\section{Cognitive LLMs: Hybrid Architectures for Human-Aligned Decision Making}

We start by providing a brief background on the central components of the ACT-R Cognitive model, before providing details about our proposed Cognitive LLM framework, \ourapproach. Our approach demonstrates essential characteristics, derived from ACT-R cognitive model, which are crucial for augmenting decision-making using foundation models with cognitive reasoning.

\subsection{Cognitive Architectures based on ACT-R}

\ourapproach~relies on an ACT-R cognitive model capable of (1) executing tasks from problem definition using decision-making behaviors observed in humans and retrieving knowledge representations similarly, (2) integrating personas ranging from novice to intermediate and expert levels, and (3) simulating the reinforcement learning processes of decision-makers as they transition from novice to expert.

\subsubsection{Human-aligned Cognitive Models}

We released VSM-ACTR 1.0 (see \sref{sec:app:vsm_actr}; \citet{vsm_actr}), which is a rule-based ACT-R cognitive decision-making model for manufacturing decision-making that implements multiple problem-solving strategies, through a combination of production rules.
We now provide a new version of this model, VSM-ACTR 2.0, which has incorporated the meta-cognitive processes that reflect on and evaluate the progress of chosen strategies---with an emphasis on headcount cost evaluation, through a reward structure that enables a process akin to reinforcement learning. This system allows the model to dynamically assess the impact of headcount costs on decision-making outcomes, computing a reward or penalty for each decision cycle. These rewards or penalties then propagate back to the initial production rule that initiated the decision cycle, thereby dynamically adjusting the utility of each decision-making strategy.

VSM-ACTR 2.0 integrates the prototypical decision process with insights into how cognitive models represent different levels of expertise \cite{BlessingAnderson1996, Martin2004}, categorizing users into three levels of expertise: novices, intermediates, and experts. Novices engage in decision-making using intuitive deliberative chunks. Intermediates can manage key metrics such as CT and OEE but struggle with the systematic analysis of intertwined variables. Experts, on the other hand, make judgments systematically. The cognitive model employs three types of knowledge chunks: decisions, decision merits, and goals. The `decision chunk' encodes eight slots including reduction time (goal), decision-making state (novice, intermediate, expert), and related variables. The `decision merits chunk' holds information on sector weights, defect increases by sector, and comparative defect rate increases. The `goal chunk' captures the initial production conditions and the ultimate goal of achieving the optimal decision. In addition, the model uses 18 procedural rules driven by goal-focused objectives across 20 states, covering actions such as choosing strategies, actions, working memory management, decisions, and evaluations. 

\subsubsection{Reinforcement Mechanisms in Production Systems}
Dopaminergic signals are believed to transmit reinforcement information to the corpus striatum \cite{schoenberg2007reinforcement}, traditionally signaling reward-related activities. However, these signals are now understood to represent the error signal in the temporal difference (TD) algorithm from reinforcement learning \cite{SuttonBarto1999}, which is applied in ACT-R's learning mechanism. As expressed in Eqn. 1, Each production rule in the ACT-R model has a utility---a value or strength---associated with it, which is updated using the TD algorithm: %
\[Eqn. \,1: \,
U_i(n) = U_i(n-1) + \alpha \left[ R_i(n) + U_i(n-1) \right],
\] %
where \( U_i(n) \) represents the value or utility of some item \( i \) (e.g., a production) after its \( n \)-th occurrence, and \( R_i(n) \) represents the reward received on the \( n \)-th occurrence. The parameter \( \alpha \) (\(0 < \alpha < 1\)) controls the learning rate.

A key strength of the TD algorithm is its ability to propagate rewards back to earlier critical productions, through a chain of productions, influencing their utilities. This mechanism is tied to the widely-used `softmax' function, which is also integral to ACT-R's production selection, as expressed in Equation 2. After propagation, if multiple productions compete with expected utility values \( U_j \), the probability of  of selecting production \( i \) is given by: %
\[Eqn. \,2: \,
\text{Probability}(i) = \frac{e^{U_i / \sqrt{2s}}}{\sum_{j} e^{U_j / \sqrt{2s}}},
\]
where the summation over \( j \) is over all the productions that currently have their conditions satisfied; and \( s \) is the noise.

To understand the dynamics of the learning mechanism, consider a scenario involving penalties within a decision-making process. The reward function \( R(s, f(x)) \) calculates the reward at the end of one decision-making round. This function takes two parameters: \( S \), representing the strategy used, and \( f(x) \), the outcome of the cost analysis, resulting in either a reward or a penalty. In one decision round, a penalty of -2 is computed due to the use of a novice strategy coupled with an inefficient cost. Factoring in the memory retention effect after a 0.5 time step, the subsequent penalty calculation modifies the impact of the decision:%
\[
R(S, x) - 0.5 \, \text{time-steps} = -2.5
\]%
\[
U(7) = U(6) + \alpha \left[ R(S, x) - 0.5 \, \text{time step} + U(6) \right] = -1.02
\]%
\( U(7) \) represents the utility of novice strategy production at the seventh occurrence of firing. While \( U(6) \) represents the utility at sixth occurrences; \( \alpha \) is set at 0.2, based on the learning rule from \citet{wagner1972inhibition}. This framework allows penalties to retroactively influence previous decisions, thus shaping the model's strategic choices in subsequent rounds.

\begin{figure*}[ht]
\centering
\includegraphics[width=0.9\textwidth]{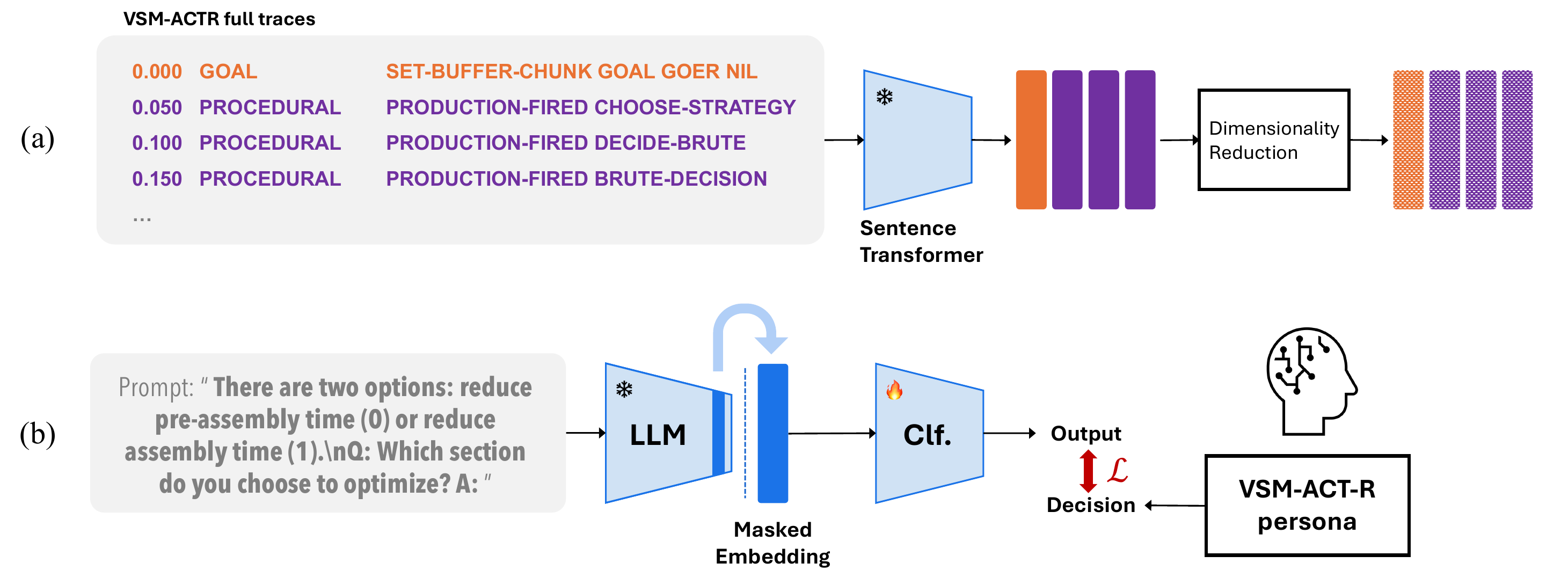}
\caption{(a) Obtaining decision representations from VSM-ACT-R. (b) LLM feature extraction for behavior prediction.}
\label{fig:components}
\end{figure*}



\subsubsection{VSM-ACTR model evaluation} 
We ran the VSM-ACTR model across 2012 decision-making trials and 32 problem sets to analyze its behavior \cite{ritterSQK2011}. Each model run comprised 15-16 trials until reach a more stable expert behavior. We encoded decision types as 0, 1, and 2 for novice, intermediate, and expert strategies, respectively.

To assess learning, individual differences, and progression, we initially used descriptive statistics to chart the average progression of decision types over 16 trials. We then employed a mixed linear model to evaluate the influence of trial numbers on decision types, incorporating repeated measures and random effects to account for individual variance. Additionally, an ordered logistic regression analyzed the relationship between the number of trials and the learning progression from novice to expert.

The results of the descriptive statistics demonstrate a significant positive impact of trial exposure on decision-making progression, evidenced by a coefficient of 0.086 ($P < 0.05$). A mixed linear model regression confirms the effect of trials on decision-making and further reveals a variance of 0.007 in the random group effects. This indicates that while there are differences between groups, these differences are relatively small, suggesting that the trials themselves predominantly explain the variability in decision type.

Threshold analysis using ordered logistic regression reveals significant transition thresholds. The transition from novice to intermediate has a significant threshold of 0.88 ($P < 0.05$), indicating a challenging progression to higher decision-making skills. In contrast, the transition from intermediate to expert shows a significantly lower threshold of 0.1 (P = 0.021), suggesting it is easier to progress from intermediate to expert than from novice to intermediate.

Comparing these results to the earlier VSM-ACTR 1.0 version, it's find that the transition from intermediate to expert has become significantly more pronounced. This change is attributed to enhancements in headcount cost evaluation factors that have reinforced the progression from intermediate to expert levels.

\subsection{The \ourapproach~Framework}

Figure \ref{fig:taskprocess} illustrates the approach to creating \ourapproach, which begins with the collection of task data and documentation. The task procedures are then modeled using ACT-R, employing stochastic simulations to analyze these tasks on a large scale. After the simulation phase, the generated synthetic data is semi-automatically distilled and combined with prompt requests. This data is subsequently used to infuse into an open source LLM through fine-tuning, resulting in a type of cognitive LLM, named \ourapproach.

\paragraph{Selecting Salient Decision Information} 
The decision-making process demonstrates cognitive reasoning through VSM-ACTR's traces, which capture the reasoning steps in real-time using a concurrent protocol (see \sref{sec:app:actr_trace} in the Appendix). These traces log the cognitive operations executed by various modules at each decision point, including the activation of the goal module to drive decisions, the use of the imaginal buffer for accessing working memory, procedural memory matching and firing, and utility updating driven by reinforcement learning, along with the decision actions.

Reserving information from ACT-R model’s decision-making traces poses challenges. A single decision-making round can generate a vast number of lines of traces, each timestamped as frequently as every 5 milliseconds. Deciding which lines to select—or whether to preserve all lines—requires a balance between minimizing information loss and reducing computational costs. The rationale for choosing outputs from specific modules as reliable sources within the decision representation lies in their clear correspondence to deterministic cognitive processes. The rationale for preserving all traces involves processes of semantic embedding extraction and dimensional reduction.

The information used to augment decision-making in this study focuses on distilling macro-level cognitive processes related to executive function\cite{gilbert2008executive}, capturing the evolution of decision-making results across trials and how decisions adapt through learning and experience. Furthermore, the decision actions are categorized into strategy levels (novice, intermediate, expert), reflecting the learning phases. Neurologically, as cognitive strategies evolve from novice to expert, there is a corresponding increase in the efficiency and effectiveness of neural circuits in the prefrontal cortex and basal ganglia in humans (see paragraphs: Implementing a reinforcement-learning mechanism in a production system framework).

\paragraph{Learning an Embedding Space of Decision Traces} 

The next step involves converting the traces into tensors that the LLM can process. This study explores two approaches: one uses selected traces, and another uses full traces.

The selected traces are components distilled from macro-level cognitive processes related to executive function. This process requires human involvement to log decision results and strategy traces, which are then numerically encoded. For instance, `0' represents a decision for reduced time in preassembly section, and `1' for assembly. These data are subsequently fed into the neural network as single vectors.

In contrast, the holistic traces approach (see Figure \ref{fig:components}a) retains both macro- and micro-level cognitive processes, with the latter including metacognition\cite{nelson1994metacognition}. Metacognition involves an awareness and understanding of one’s own cognitive processes, as exhibited through model traces that demonstrate the use of the imaginal buffer for accessing working memory, procedural memory matching and firing, headcount cost analysis, and the assessment of strategy effectiveness.

The investigation begins with the transformation of full traces from VSM-ACTR, representing both cognitive and metacognitive processes, into a format that balances information retention with computational efficiency. Cognitive reasoning traces for each task are processed through a sentence transformer to obtain semantic embeddings for each timestamp. A Sum of Ranked Explanatory Effects (SREE) analysis is then applied to determine the number (N) of principal components that account for at least 70\% of the variance. Finally, these embeddings are reduced to N dimensions using Principal Components Analysis (PCA)\cite{abdi2010principal}. 


\paragraph{Injecting Decision Information into LLMs} 

With the VSM-ACTR model, which represents human-like cognitive reasoning in repeated decision-making tasks, this section outlines the experimental settings for fine-tuning of the \ourapproach~framework. Fine-tuning, sometimes referred to as transfer learning, involves optimizing all model weights for the given task. The process includes parsing consistent template prompts that reflect the decision making task into an open-source LLM, aligning the task for the cognitive model Using the LLM as the base model to access the last hidden layer and obtain masked embeddings, constructing a classification layer with softmax activation on top of the base model, using targets containing the salient decision representations of the cognitive model and features from the masked embeddings of the base LLM, and fine-tuning the LLM for classification using the LORA method (see \sref{sec:app: additional figures and illustrations}). The key points are: (1) The targets decode the salient decision information from the cognitive model. (2) Use the final layer of contextualized embeddings in transformer-based LLMs, generated through the attention block mechanism. The attention block, a key feature of transformers, distinguishes them from other architectures like recurrent neural networks \cite{graves2012sequence}. It creates embeddings that capture the in-context meaning of tokens by recombining them with other tokens' embeddings. Successive attention blocks further refine these embeddings, producing multiple layers of abstraction. The final layer, a blend of these refined embeddings, is used in this pipeline because it offers the richest semantic information while balancing minimal information loss and reduced computational costs for fine-tuning. (3) Use Low-Rank Adaptation (LoRa) for its efficiency in fine-tuning, reducing the computational resources and time required while maintaining high model performance \cite{hu2022lora}.

\section{Experiments}
\subsection{Problem Setting}

As an instantiation of the problem definition, above, our manufacturing line has two sections with potential defect sources: pre-assembly and assembly. Pre-assembly takes 40 seconds with an OEE rate of 88\%, while assembly takes 44 seconds with an OEE rate of 80.1\%. To reduce total assembly time by 4, we must identify which section can be shortened with minimal defect increase. We note that reducing cycle time will also lead to an increase in headcount costs. 

\subsection{Implementation Details}
The LlaMa-2 13B model was chosen as the foundation for this research because of its demonstrated effectiveness and efficiency in NLP tasks (Huang et al., 2024). As a state-of-the-art large language model, LlaMa has been trained on trillions of tokens from publicly available datasets. Unlike other transformer-based models such as the GPT family, which can only be accessed at the user's end, LlaMa's architecture, including its pre-trained weights, is fully accessible. Furthermore, its proven capability to extract the last hidden layer for predicting behavioral discrepancies has been provided (Binz and Schulz, 2024). These attributes collectively establish LlaMa-2 13B as an optimal choice for this study.

To determine the dataset size that can effectively perform the task while balancing efficacy and resource limitations, we referred to \citet{Kumar2024OptimalNLP}, who showed evidence that LlaMa-2 13B achieves F1 scores above 0.9 in resource-limited text classification tasks, with datasets as 1,000 rows per class. Based on this, we developed the dataset size for fine-tuning as \( N \) (number of classes) * 1,000. The ACT-R dataset for binary decision-making classification contains 2,012 decision-making trials, Obtained by running the developed ACT-R model across 32 problem sets, each ACT-R persona was run for 15-16 trials until more stable expert behavior was achieved \cite{ritterSQK2011}.

\subsection{Baseline Models} 
This study compared the goodness-of-fit and prediction accuracy of the resulting models using holdout data against two baselines: a random guess model and LlaMa without fine-tuning, obtained by reading out log-probabilities of the pre-trained LlaMa.

A random guess model serves as the most basic form of chance level baseline and represents the simplest hypothesis for model comparison. In psychological interdisciplinary experiments, control conditions often employ random responses to distinguish the effects of treatment from chance \cite{gaab2019effects}. This approach allows assessing the extent to which decisions are influenced by knowledge versus being purely stochastic.

On the other hand, using LlaMa without fine-tuning as a baseline provides a reference point to measure the impact of fine-tuning on the model’s performance. This comparison reveals how much the model `learns' from the fine-tuning process compared to its generic, pre-trained state.


\subsection{Research Questions}

Based on our framework's components, we identify a set of research questions that we answer through experiments.


\begin{enumerate}[label=RQ\arabic*:]
    \item \textit{What are the properties of a useful neural network representation of the decision-making process in Cognitive Architectures?}
\end{enumerate}
Answering this question sets the groundwork for developing a context-aware domain knowledge base for augmenting decision-making in LLMs.

\begin{enumerate}[label=RQ\arabic*:,resume]
    \item \textit{What level of complexity in behavior representation can LLMs effectively capture?}
\end{enumerate}
Previous research has used LLM conceptual embeddings to predict human behavior based on past behavioral studies \cite{BinzSchulz2023a}, confirming LLMs' ability to replicate known human patterns. However, high costs and extensive data collection efforts limit this method. By incorporating cognitive model simulations, the study seeks to address these limitations and broaden the investigation to determine the extent to which LLMs can reproduce decision-making knowledge. This will, in turn, help define the depth of decision-making domain knowledge that can be effectively integrated with the innate learning capabilities of LLMs.

\begin{enumerate}[label=RQ\arabic*:,resume]
    \item \textit{Can we inform the LLM with knowledge about the reasoning process of the cognitive architecture?}
\end{enumerate}
Inspired by previous works on knowledge-injection \cite{oltramari2021generalizable, ma2019towards}, answering this question offers insights into knowledge transfer from domain-specific bases to LLMs and evaluates its impact on performance in holdout tasks. The method for addressing RQ1 was introduced in the first two sections of our approach framework.

\subsection{Feature Extraction for Behavior Prediction}

To answer RQ2: What level of complexity in behavior representation can LLMs effectively capture? Building on previous research that used conceptual embeddings from LLMs to predict human behavior with historical behavioral data \cite{BinzSchulz2023a}, we adopted the same method of LLM feature extraction for behavior prediction \cite{hussain2023tutorial}. We created datasets consisting of last contextual embeddings as features and the corresponding different levels of VSM-ACTR decision actions representations as targets. We obtained embeddings by passing prompts that included all the information that VSM-ACTR had access to on a given trial through LlaMa and then extracting the hidden activations of the final layer, as shown in Figure \ref{fig:components}b.

The first dataset used features extracted from prompts (see \sref{sec:app: prompt templates}) identical to the VSM-ACTR task, with targets being the VSM-ACTR decision-making results, where '0' indicates reduced time in preassembly and '1' indicates assembly. The second dataset's prompt template added an explanation of the strategy adopted by VSM-ACTR and used compound targets comprising both the decision-making results and the strategies reflecting the learning trajectory (novice, intermediate, and expert). The targets were encoded as follows: 0, 1, and 2 for preassembly choices using novice, intermediate, and expert strategies, respectively, and 3, 4, and 5 for assembly choices following the same pattern. With these two datasets, we fitted a regularized logistic regression model using 10-fold cross-validation for dataset 1 and multinomial regression using 10-fold cross-validation with L2 regularization for dataset 2. Model performance was assessed by measuring the goodness of fit through negative log-likelihood (NLL) and the predictive accuracy of hold-out data.

\subsection{Fine Tuning for Knowledge Transfer}To answer RQ3: whether LLMs can be informed with knowledge about the reasoning processes of cognitive architecture—we use the fine-tuning approach of \ourapproach~Framework. The fine-tuning process employs Cross-Entropy as the loss function and uses Adam optimization. Training involves a train test split of 0.2 and uses a batch size of 5 for both training and validation phases. The learning rate is set to 1e-5, with the training spanning across 10 epochs. To ensure regularization and prevent overfitting, a weight decay of 0.01, and a dropout of 0.5 are applied, and gradient accumulation is set to 2. Last but not least, gradient clipping is employed to maintain a maximum gradient norm of 1.0 for gradient explosion control. %
We evaluate the model fitting and generalization quality using training loss and validation loss across epochs, then compare the goodness of fit and prediction accuracy of the hold-out data against the baseline models.

\section{Results}

\subsection{Finding Useful Decision Process Embeddings}
 The approach of distilling macro-level cognitive processes related to executive function captures the evolution of decision-making results across trials and how decisions adapt through learning and experience, all represented as a sequential single vector. This format facilitates ease of use for downstream tasks involving knowledge transfer. However, this method retains only partial cognitive decision-making knowledge.

In contrast, the holistic semantic preservation approach encompasses both macro and micro-level cognition processes. However, the embeddings produced vary in shape due to the individual differences in traces originating from stochastic simulations. They cannot be directly fed into neural networks for downstream tasks. Nevertheless, the first two principal components of the reduced embeddings, which correspond to the semantic mapping of ACT-R's components—including procedural, imaginal, goal knowledge, utility updating, and decision-making—are detailed in Figure 6.

The MANOVA analysis was conducted to assess the overall effect of the independent variables, which include label categories or ACT-R components, on the combined dependent variables—components of reduced embeddings. This analysis reveals a significant relationship with the semantic mapping of ACT-R's components. For instance, the extremely low Wilks' lambda value (0.0004) suggests that the label or ACT-R component categories explain nearly all the variance in the dependent variables, indicative of a strong group effect. The statistical tests applied—Wilks' lambda, Pillai's trace, Hotelling-Lawley trace, and Roy's greatest root—all demonstrate strong significance, as evidenced by the extremely low p-values across all tests. These findings highlight that the principal components retained in the PCA successfully capture the essential variance related to these cognitive processes.

This result validates that our semantic abstraction method has the potential to retain the maximum semantics of neural symbolic representations at a minimal computational cost. However, further work is required to address the issue of ragged tensors for downstream tasks.

In a preliminary experiment, we addressed the issue of ragged tensors by employing padding with value imputation. We then integrated the 240 full cognitive reasoning traces from the VSM-ACTR model with LLM using embedding concatenation and conducted feature extraction for behavior prediction. Specifically, we transposed the reduced embeddings from each cognitive model run into a (1, X) dimension tensor and subsequently concatenated this with the LLM’s last contextual embedding from the same prompt. These concatenated embeddings served as resources for predicting decision-making within the VSM-ACTR model. The prediction targets were multifaceted, including both the decision-making results and the strategies used. The results showed no significant improvement in prediction accuracy with concatenated embeddings compared to using LLM embeddings alone. Further details, illustrations, discussions on potential reasons, and suggestions for improvements, are provided in \sref{sec:app:additional_experiments}, in the Appendix.

\begin{figure}[ht]
\centering
\includegraphics[width=\columnwidth]{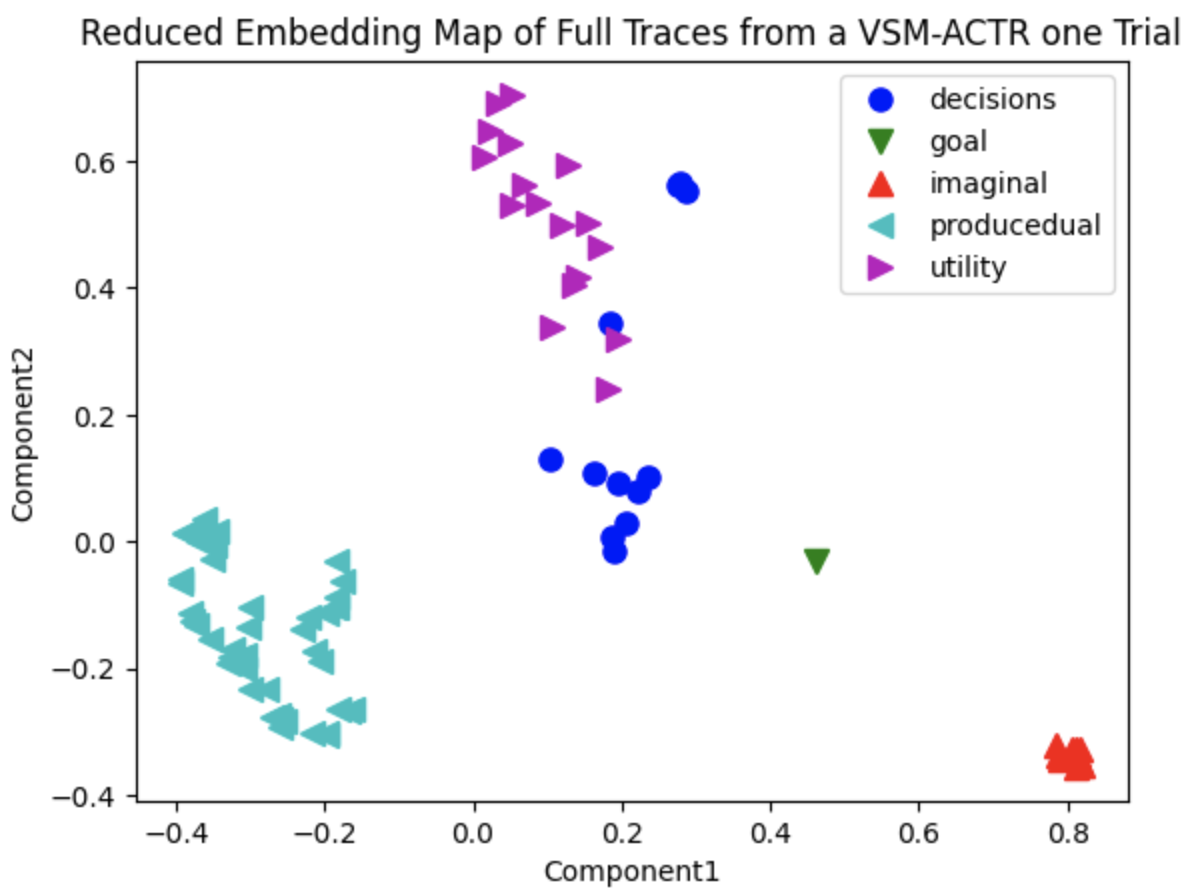}

\caption{ACTR embedding mapping}
\label{fig:actr_embedding}
\end{figure}

\subsection{Assessing Behavior Complexity Captured by LLMs} 
Table \ref{tab:metrics} shows that \ourapproach~captures a single facet of decision-making, achieving an average accuracy of 0.64 across 10 validation folds in the holdout task. When decision-making targets involve multiple facets—encompassing both choices and strategies that shape the learning trajectory—the accuracy decreases to 0.42. While this reduction suggests that capturing complex decision-making processes is less accurate, the results still show promise in handling these complexities. However, the Negative Log-Likelihood (NLL) reveals greater predictive uncertainty for multifaceted decision-making processes, as evidenced by a significantly higher NLL of 1.18 compared to 0.65 in single-facet scenarios.

\begin{table}[htbp]
\centering
\caption{Evaluation for Single and Multi Facets Targets}
\begin{tabular}{@{}lcc@{}}
\toprule
\textbf{Target Type} & \textbf{NLL} & \textbf{Accuracy} \\ \midrule
Single Facet Target  & 0.63         & 0.64              \\
Multi Facets Target  & 1.18         & 0.42              \\
\bottomrule
\end{tabular}
\label{tab:metrics}
\end{table}

\begin{table}[h]
\vspace{-0.5cm}
\centering
\caption{Comparison of VSM-ACTR with baselines}
\begin{tabular}{lcc}
\toprule
\textbf{Model} & \textbf{NLL} & \textbf{Accuracy} \\ \midrule
Chance-level& 0.6931 & 0.4826 \\
LlaMa & 1.1330  & 0.3564\\
\ourapproach~(ours)& \textbf{0.6534} & \textbf{0.6576} \\
\bottomrule
\end{tabular}
\label{tab:accuracy_nll_comparison}
\end{table}

\subsection{Injecting LLMs with CA Decision Process}
We first report training and validation losses, across 10 epochs, to reveal the fine-tuned model's learning and generalization behavior. %
Initially, the training loss begins at approximately 0.73, with a slight fluctuation observed in subsequent epochs, peaking around epoch 2 and showing a notable dip at epoch 7. In contrast, the validation loss starts at around 0.64 and remains remarkably stable throughout the epochs. This consistency in validation loss, coupled with a generally downward trend in training loss after its initial variations, suggests that the model is learning effectively. The overall trend indicates an improvement in model performance over time, reflecting its capability to generalize well on unseen data.

We then report the comparison of the \ourapproach~with the baseline models on goodness of fit using negative log likelihood (NLL) and accuracy score for hold-out data. The \ourapproach~model demonstrates significantly better performance across all metrics compared to the LlaMa-only model, highlighting its effectiveness in decision-making tasks involving sequential cognitive reasoning. Additionally, the LlaMa-only model performs worse than the chance-level model. This underscores the necessity of fine-tuning pre-trained language models like LlaMa to adapt them to specific human-aligned repeated decision-making tasks.

\section{Discussion and Conclusion}
\label{sec:discussion}

\paragraph{Main Insights/Takeaways} 
Resolving the dichotomy between the human-like yet constrained reasoning processes of CAs and the broad, often noisy inference behavior of LLMs remains a challenging but exciting pursuit. This is crucial for enabling reliable machine reasoning capabilities in production systems. This study introduces \ourapproach, a novel neuro-symbolic architecture designed to enhance human-aligned and versatile decision-making by integrating the ACT-R model's cognitive process with LLMs. Our framework extracts and embeds ACT-R model's internal decision-making processes as latent neural representations based on using traces of its performance, then injects this information into trainable LLM adapter layers, and finally fine-tunes the LLMs for downstream prediction tasks. \ourapproach~addresses the data scarcity issue often encountered in research aimed at aligning LLMs with human reasoning. Our approach demonstrates improved grounded decision-making capabilities compared to LLM-only baselines that leverage chain-of-thought reasoning strategies.

We explore distilling latent representations. The findings show that distilling macro-level cognitive processes preserves high-level neural symbolic knowledge, aiding downstream tasks but only partially capturing decision-making knowledge. A holistic semantic preservation approach, covering both cognitive and metacognitive processes, better retains full neural symbolic semantics with low computational costs. However, challenges with ragged tensors in downstream tasks require further research. We then use a VSM-ACTR cognitive model, developed for a manufacturing design task, to distill its macro-level cognitive processes as domain knowledge. This knowledge was then employed in both a feature extraction for behavior prediction method and a fine-tuning pipeline to investigate the LLM's capabilities in (1) capturing the complexity of behavioral representations and (2) determining whether and how the LLM can be informed by the reasoning processes inherent in the cognitive architecture. The results show that (1) prompt embeddings generated through feature extraction capture repeated human-aligned decision-making and the overall structure of learning. However, they struggle to capture complex decision-making rationales, such as the strategies employed. (2) Fine-tuning effectively informs the LLM with knowledge of decision-making tasks involving sequential cognitive reasoning.

\paragraph{Limitations} 
The limitation lies in achieving close human alignment with the VSM-ACTR model. While the model is currently tuned to reflect general human learning behavior patterns in decision-making, it still requires more granular human data for cognitive fine-tuning. The closer the VSM-ACTR model aligns with human behavior, the more accurately it can represent human decision-making processes.

\paragraph{Future Work} 
To address this limitation, the project aims to enhance the human alignment of the cognitive model by fine-tuning it with granular human data. This will be followed by integrating the full semantic space of ACT-R into the LLM's decision-making processes. The strategy of using semantic abstraction from cognitive model trace lines demonstrates the potential. However, the preliminary experimental results highlight challenges in integrating such knowledge, which are worth further exploration. Grounding LLMs in a robust cognitive framework while minimizing computational costs could pave the way for effectively scaling the infusion of superior domain knowledge into LLMs.

\bibliography{aaai25}

\clearpage
\onecolumn
\section{Appendix}

\subsection{Additional Figures and Illustrations}
\label{sec:app: additional figures and illustrations}
\begin{figure}[!h]
\centering
\includegraphics[width=\columnwidth]{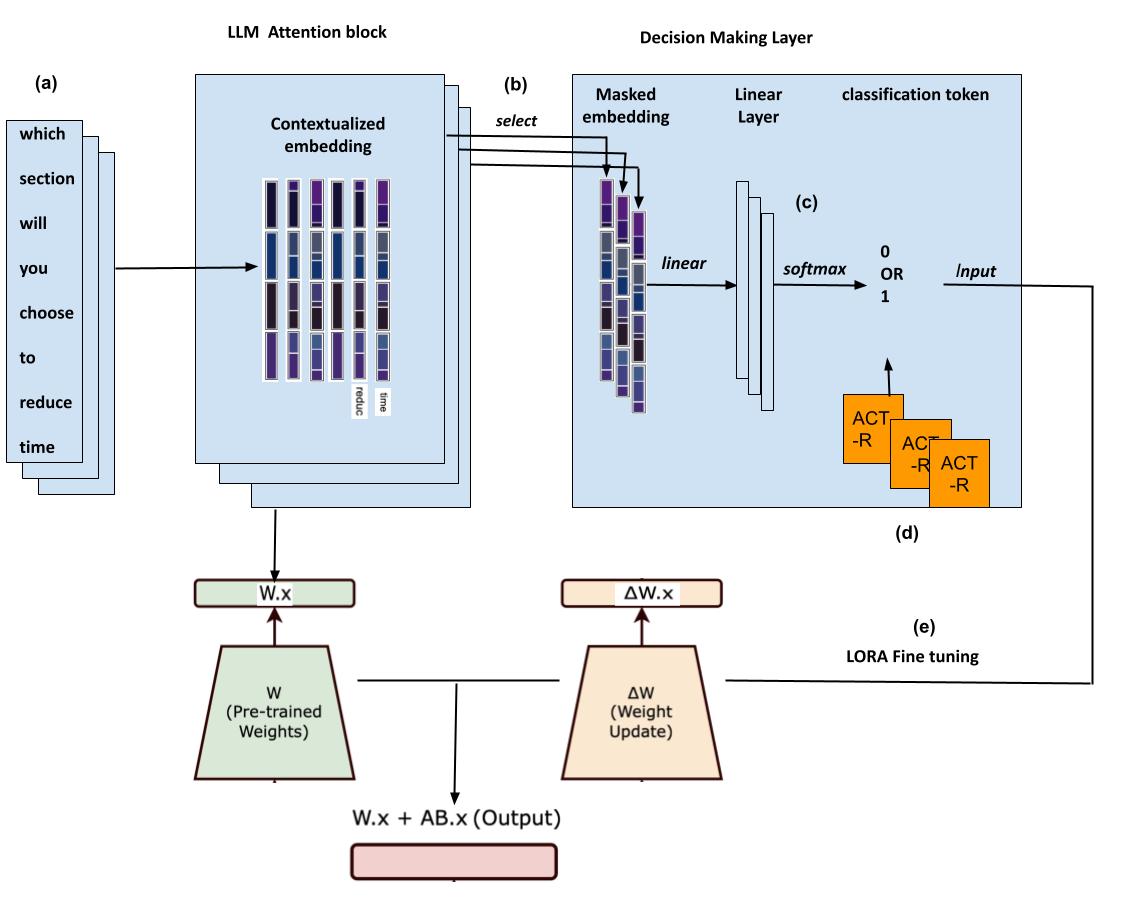}

\caption{Finetuning pipeline.} 
\label{fig:finetuning}
\end{figure}
The fine-tuning begins by (a) parsing consistent template prompts that reflect the decision making task into an open-source LLM, aligning the task for the cognitive model; (b) using the LLM as the base model to access the last hidden layer and obtain masked embeddings; (c) constructing a classification layer with softmax activation on top of the base model; (d) using targets containing the salient decision representations of the cognitive model and features from the masked embeddings of the base LLM; (e) and fine-tuning the LLM for classification using the LORA method

\subsection{Additional Experiments}
\label{sec:app:additional_experiments}

\begin{figure}[!h]
\centering
\includegraphics[width=\textwidth]{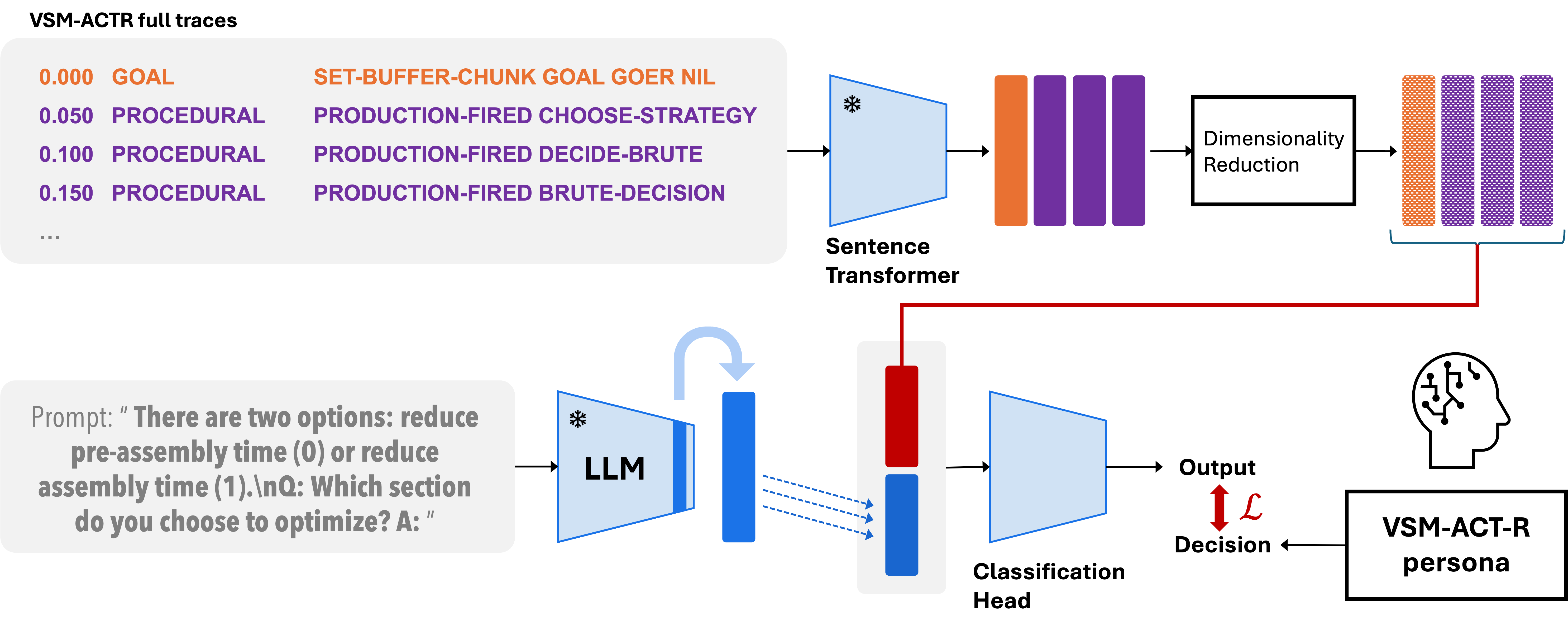}
\caption{Full knowledge-extraction and feature-injection pipeline.}
\label{fig:app:full_injection}
\end{figure}

\paragraph{Extended feature-injection pipeline} Here, we show the preliminary results of combining masked embeddings, extracted from open-source LLMs (e.g., LlaMa), with embeddings of ACT-R's internal decision-making process for training a prediction head on top --- as illustrated in Figure \ref{fig:app:full_injection}. We integrated the 240 full cognitive reasoning traces from the VSM-ACTR model with LlaMa using embedding concatenation and conducted feature extraction for behavior prediction. Specifically, we transposed the reduced embeddings from each cognitive model run into a (1, X) dimension tensor and subsequently concatenated this with the LlaMa’s last contextual embedding from the same prompt. These concatenated embeddings served as resources for predicting decision-making within the VSM-ACTR model. The prediction targets were multifaceted, including both the decision-making results and the strategies reflecting the learning trajectory (novice, intermediate, and expert). 

We employed a train-test split of 0.4 and conducted multinomial regression with L2 regularization, using two-fold cross-validation to compare the prediction accuracy and goodness of fit, as measured by negative log-likelihood, between concatenated embeddings and LlaMa embeddings alone. The results indicated that the prediction accuracy for both datasets was 0.71, with the concatenated embeddings showing a slightly better negative log-likelihood of 0.9535 compared to 0.9553 for LlaMa-only embeddings.

The current findings suggest no significant improvement in behavior prediction when combining masked embeddings versus using LlaMa embeddings alone. One possible explanation is the relative scale of the VSM-ACTR reduced embeddings compared to those of LlaMa, which is disproportionately small (1:10). Consequently, the LlaMa embeddings may dominated the decision-making process within the model due to their larger scale. A potential solution could be to generate longer VSM-ACTR model traces, including tenfold more decision-making trials, thereby enhancing the scale and variability of its features.

Also, the method we use to handle ragged tensors—padding followed by value imputation—could potentially dilute the VSM-ACTR embeddings and reduce their accuracy. Finding an alternative method to preserve the full embeddings from VSM-ACTR may potentially improve the results.

Lastly, the limited dataset size could be influencing the results. The preliminary test used only 240 complete traces. Expanding the dataset may provide more insights into the performance of the proposed approach.

\subsection{Description of the Code and Data}

In LLM-ACTR-main.zip

\subsection{Background: the \texttt{VSM-ACTR 1.0} Model}
\label{sec:app:vsm_actr}

{Here is a brief description of VSM-ACTR \cite{vsm_actr} --- to support the discussion in the main content of this paper. It is a concurrent work that the present manuscript extends. This referenced manuscript has been peer-reviewed and has been accepted for publication in an upcoming conference venue.}

This section starts with capturing intuition and domain knowledge from decision makers, followed by the model structure and learning mechanism, and concludes by examining a model output snippet from one run of our VSM model.

\subsubsection{Model Design}
The model, built upon the prototypical decision process distilled by Bosch plant managers, incorporates how cognitive models are designed for different levels of expertise. For novices, the model utilizes intuitive deliberative chunks to make decisions. For intermediates, it understands key metrics such as cycle time (CT) and Overall Equipment Effectiveness (OEE). However, intermediates often lack the ability to systematically analyze how these metrics interrelate and cumulatively impact efficiency and quality. Experts, on the other hand, make well-informed judgments based on a comprehensive view of all relevant metrics, obtained through Value Stream Mapping (VSM).
\subsubsection{Declarative Chunks}
We created chunks representing knowledge from intuitions to professional expertise. These representations are divided into three chunk types: decisions, decision merits, and goals. Decision chunk encodes six slots: reduction time, decision-making state (e.g., novice, intermediate, expert), OEE, and CT. The decision merits chunk holds knowledge on weights for sectors, defect increase for sectors, and the difference in defect rate increase between the two. The goal chunk encodes the initial production conditions and the ultimate goal of making the optimal decision.

\subsubsection{Production Rules}
Three sets of production rules represent the decision-making behaviors of novice, intermediate, and expert decision-makers. These sets comprise a total of 17 rules, each driven by goal-focused objectives across 14 states.

We use the expert production rule set as an example, as shown in Figure 2. Once the decision-choice center decides to activate this set of expert decision productions, it starts by perceiving the problem and retrieving related decision-making metrics from chunks. The imaginal buffer then acts as a temporary workspace, holding and manipulating relevant information during decision-making. It allows the model to build new mental representations or modify existing ones based on incoming data or problem-solving needs. This involves using the imaginal buffer to assess the relationships between the decision target and decision metrics, particularly considering the impact of each sector's weight on the defect rate change, and determining the final defect rate increase for each sector. These results are stored in the imaginal buffer and later retrieved for comparison. This then allows the model to select the sector with the lowest defect increase.
\begin{figure}
\centering
\includegraphics[width=1\textwidth]{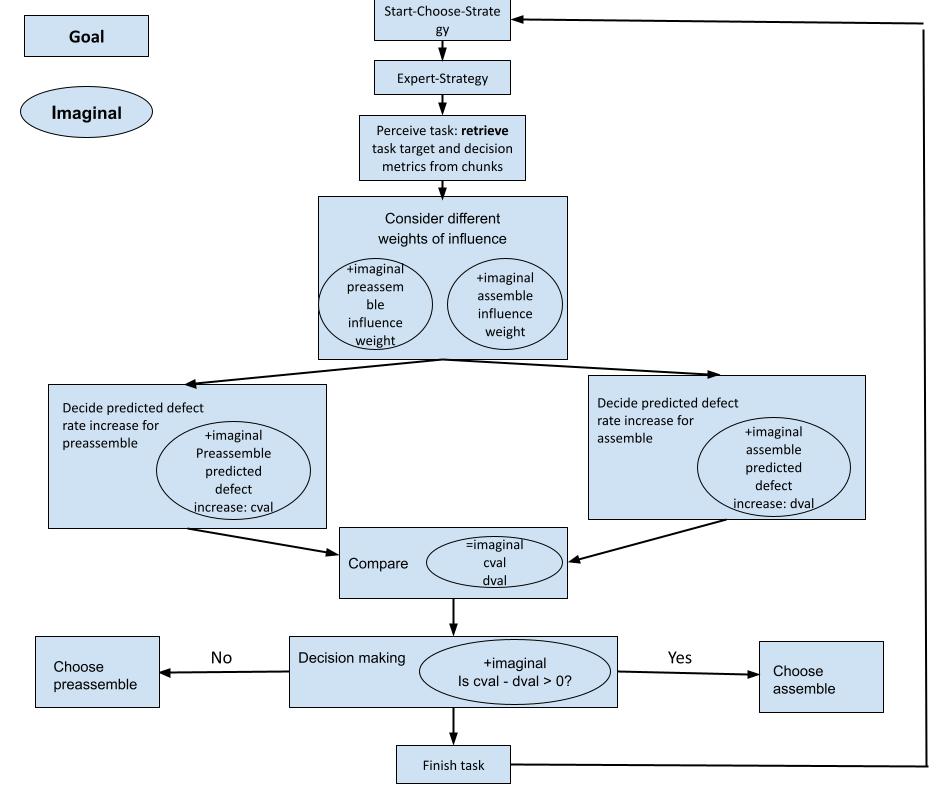}
\caption{Production rules control structure for expert decision making and their use of the ACT-R Goal and Imaginal buffers} \label{fig1}

\end{figure}

\subsubsection{Level of Expertise Mechanism}
The model can learn while performing tasks through two mechanisms leading to varying levels of expertise, as shown in Figure 3.

\begin{figure}[ht]
\centering
\includegraphics[width=1\textwidth]{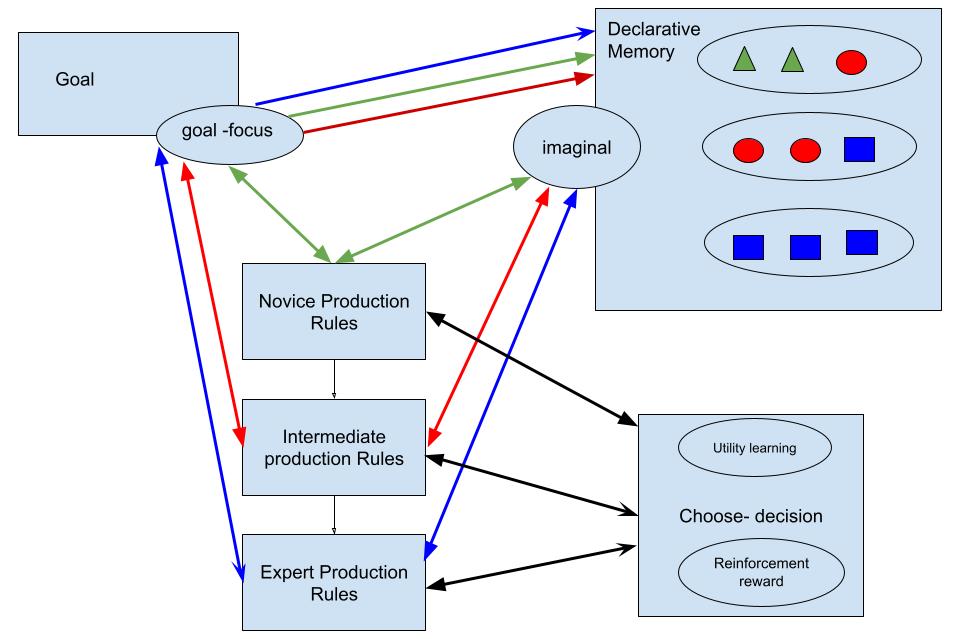}

\caption{Level of expertise mechanism in VSM-ACT-R}
\label{fig:expertise}
\end{figure}
The model mimics human decision-making behavior through differentiating knowledge representations.
\textbf{Declarative Memories}: These memories store knowledge that aligns with human intuition and expertise gained from the VSM. For example, the green triangles in the figure represents a portion of the intuition used by novice decision-makers.
\textbf{Production Rules}: These rules capture the rational decision-making processes observed in human subjects. The green (lighter) lines illustrate how the imaginal buffer retrieves relevant portions of the novice declarative memory and feeds them to the novice production rule set.
Intermediate and expert decision-making levels follow the same principle. Red and blue shapes represent their respective declarative memory chunks, and the corresponding (darker) colored arrows show the flow of information through their production rule sets. Finally, the goal buffer utilizes the "goal focus" command to manipulate the different phases of the task.

Beyond mimicking human behavior, the model also simulates the learning progress achieved by the \textbf{Decision-Choice Control}, which manages errors, learning, and memory through utility learning and reinforcement rewards. Novice decision-making starts with a utility base and includes a noise setting. The intermediate and expert production rules receive rewards when the corresponding decision-making results are achieved. The utility of these production rules updates is based on the rewards received and the retention of memory, which depends on the time passed since the rule last fired.

\subsection{Example ACT-R Decision Trace}
\label{sec:app:actr_trace}
We provide a reference ACT-R trace, which we explain is the basis for extracting the decision-making knowledge representation.

The model begins by setting up the goal (line 1), followed by starting with a novice strategy (line 3). For the production rules associated with each strategy, each production rule's utility is updated based on the reward received and the time since the last selection. For example, the utility of the NAIVE-CHOICE rule decreased from 3 to 1.96 (lines 14-16) due to a penalty of -2.25 for the time passed since the last selection. As the utility of naive strategies decreases, the likelihood of EXPERT-Strategy (lines 87-89) being triggered increases.\\

\lstset{
    basicstyle=\ttfamily,
    keywordstyle=\color{blue}\bfseries,
    commentstyle=\color{gray}\itshape,
    stringstyle=\color{green},
    showstringspaces=false,
    breaklines=true,
    frame=single,
    columns=fullflexible,
    numberstyle=\tiny\color{codegray},
}

\begin{lstlisting}
0.000   GOAL                   SET-BUFFER-CHUNK GOAL GOER NIL
0.050   PROCEDURAL             PRODUCTION-FIRED CHOOSE-STRATEGY
0.100   PROCEDURAL             PRODUCTION-FIRED DECIDE-BRUTE
0.150   PROCEDURAL             PRODUCTION-FIRED BRUTE-DECISION
assembly is always a good place to reduce time!
0.200   PROCEDURAL             PRODUCTION-FIRED REHEADCOUNT
-0.01999998 
0.250   PROCEDURAL             PRODUCTION-FIRED STOP
this is the end of one decision making
Utility updates with Reward = -2.0   alpha = 0.2
Updating utility of production CHOOSE-STRATEGY
U(n-1) = 0.0   R(n) = -2.25 [-2.0 - 0.25 seconds since selection]
U(n) = -0.45000002
Updating utility of production DECIDE-BRUTE
U(n-1) = 3.0   R(n) = -2.2 [-2.0 - 0.2 seconds since selection]
U(n) = 1.96
Updating utility of production BRUTE-DECISION
U(n-1) = 0.0   R(n) = -2.15 [-2.0 - 0.15 seconds since selection]
U(n) = -0.43000004
Updating utility of production REHEADCOUNT
U(n-1) = 0.0   R(n) = -2.1 [-2.0 - 0.1 seconds since selection]
U(n) = -0.42
Updating utility of production STOP
U(n-1) = 0.0   R(n) = -2.05 [-2.0 - 0.05 seconds since selection]
U(n) = -0.41
0.300   PROCEDURAL             PRODUCTION-FIRED CHOOSE-STRATEGY
0.350   PROCEDURAL             PRODUCTION-FIRED DECIDE-INTERMEDIATE
0.400   PROCEDURAL             PRODUCTION-FIRED INTERMEDIATE-STRATEGY
0.01999998 
0.600   IMAGINAL               SET-BUFFER-CHUNK-FROM-SPEC IMAGINAL 
0.650   PROCEDURAL             PRODUCTION-FIRED INERMEDIATE-CHOICE2
choose assemble has better stable output!
0.700   PROCEDURAL             PRODUCTION-FIRED REHEADCOUNT
-0.01999998 
0.750   PROCEDURAL             PRODUCTION-FIRED STOP
this is the end of one decision making
Utility updates with Reward = 0.0   alpha = 0.2
Updating utility of production CHOOSE-STRATEGY
U(n-1) = -0.45000002   R(n) = -0.5 [0.0 - 0.5 seconds since selection]
U(n) = -0.46
Updating utility of production DECIDE-INTERMEDIATE
U(n-1) = 0.0   R(n) = -0.45 [0.0 - 0.45 seconds since selection]
U(n) = -0.089999996
Updating utility of production INTERMEDIATE-STRATEGY
U(n-1) = 0.0   R(n) = -0.4 [0.0 - 0.4 seconds since selection]
U(n) = -0.080000006
Updating utility of production INERMEDIATE-CHOICE2
U(n-1) = 0.0   R(n) = -0.15 [0.0 - 0.15 seconds since selection]
U(n) = -0.030000001
Updating utility of production REHEADCOUNT
U(n-1) = -0.42   R(n) = -0.1 [0.0 - 0.1 seconds since selection]
U(n) = -0.35599998
Updating utility of production STOP
U(n-1) = -0.41   R(n) = -0.05 [0.0 - 0.05 seconds since selection]
U(n) = -0.338
0.800   PROCEDURAL             PRODUCTION-FIRED CHOOSE-STRATEGY
0.850   PROCEDURAL             PRODUCTION-FIRED EXPERT-STRATEGY
0.900   PROCEDURAL             PRODUCTION-FIRED PERCEIVE
0.950   PROCEDURAL             PRODUCTION-FIRED PREASSEMBLE-WEIGHT
0.4836186 
caculate the preassemble defect decision weight
1.150   IMAGINAL               SET-BUFFER-CHUNK-FROM-SPEC IMAGINAL 
1.200   PROCEDURAL             PRODUCTION-FIRED ASSEMBLE-WEIGHT
0.5163814 
calculate the assemble defect decision weight
1.400   IMAGINAL               SET-BUFFER-CHUNK-FROM-SPEC IMAGINAL 
1.450   PROCEDURAL             PRODUCTION-FIRED PREASSEMBLE
0.23213693 
calculate the final preassemble defect rate
1.650   IMAGINAL               SET-BUFFER-CHUNK-FROM-SPEC IMAGINAL 
1.700   PROCEDURAL             PRODUCTION-FIRED ASSEMBLE
0.28917354 
calclate the assemble defect rate
1.900   IMAGINAL               SET-BUFFER-CHUNK-FROM-SPEC IMAGINAL 
1.950   PROCEDURAL             PRODUCTION-FIRED COMPARE
-0.05703661 
2.000   PROCEDURAL             PRODUCTION-FIRED DECIDE
choose preassemble has better stable output!
2.050   PROCEDURAL             PRODUCTION-FIRED HEADCOUNT
0.01999998 
2.100   PROCEDURAL             PRODUCTION-FIRED STOP
this is the end of one decision making
Utility updates with Reward = 6.0   alpha = 0.2
Updating utility of production CHOOSE-STRATEGY
U(n-1) = -0.46   R(n) = 4.65 [6.0 - 1.35 seconds since selection]
U(n) = 0.56200004
Updating utility of production EXPERT-STRATEGY
U(n-1) = 0.0   R(n) = 4.7 [6.0 - 1.3 seconds since selection]
U(n) = 0.94
Updating utility of production PERCEIVE
U(n-1) = 0.0   R(n) = 4.75 [6.0 - 1.25 seconds since selection]
U(n) = 0.95
Updating utility of production PREASSEMBLE-WEIGHT
U(n-1) = 0.0   R(n) = 4.8 [6.0 - 1.2 seconds since selection]
U(n) = 0.96000004
Updating utility of production ASSEMBLE-WEIGHT
U(n-1) = 0.0   R(n) = 5.05 [6.0 - 0.95 seconds since selection]
U(n) = 1.0100001
Updating utility of production PREASSEMBLE
U(n-1) = 0.0   R(n) = 5.3 [6.0 - 0.7 seconds since selection]
U(n) = 1.0600001
Updating utility of production ASSEMBLE
U(n-1) = 0.0   R(n) = 5.55 [6.0 - 0.45 seconds since selection]
U(n) = 1.11
Updating utility of production COMPARE
U(n-1) = 0.0   R(n) = 5.8 [6.0 - 0.2 seconds since selection]
U(n) = 1.1600001
Updating utility of production DECIDE
U(n-1) = 0.0   R(n) = 5.85 [6.0 - 0.15 seconds since selection]
U(n) = 1.17
Updating utility of production HEADCOUNT
U(n-1) = 0.0   R(n) = 5.9 [6.0 - 0.1 seconds since selection]
U(n) = 1.1800001
Updating utility of production STOP
U(n-1) = -0.338   R(n) = 5.95 [6.0 - 0.05 seconds since selection]
U(n) = 0.91959995
\end{lstlisting}

\lstdefinestyle{prompt}{
    basicstyle=\ttfamily\footnotesize,
    keepspaces=true, 
    showspaces=false,
    showstringspaces=false,
    showtabs=false,
    keywordstyle=\color{blue}\bfseries,
    commentstyle=\color{gray}\itshape,
    stringstyle=\color{green},
    breaklines=true,
    tabsize=2
    frame=single,
    columns=fullflexible,
    numbers=none,
    literate={\\\%}{\%}1,
    escapechar=\%
}

\lstdefinestyle{response}{
    basicstyle=\ttfamily\footnotesize,
    backgroundcolor=\color{backcolour}, 
    keywordstyle=\color{blue}\bfseries,
    commentstyle=\color{gray}\itshape,
    stringstyle=\color{green},
    showstringspaces=false,
    breaklines=true,
    frame=single,
    columns=fullflexible,
    numbers=none,
    literate={\\\%}{\%}1,
    escapechar=\%
}

\subsection{LLM System Prompt Templates}
\label{sec:app: prompt templates}

\begin{llmprompt}{gray}{Prompt template for fine-tuning and single-facet target behavior prediction:}
Our manufacturing line has two sections with potential defect sources: pre-assembly (0) and assembly (1). Pre-assembly takes {CT1} seconds with an Overall Equipment Effectiveness(OEE) rate of \{OEE1\}\%, while assembly takes \{CT2\} seconds with an OEE rate of \{OEE2\}\%. To reduce total assembly time by 4 seconds, we need to identify which section can be shortened with minimal defect increase. It's important to note that reducing cycle time will also lead to an increase in line headcount costs. There are two options: reduce pre-assembly time (0) or reduce assembly time (1).\\ \\

\noindent Question: Which section do you choose to optimize? \\ \\

\noindent Answer: 
\end{llmprompt}

\begin{llmprompt}{gray}{Prompt template for multi-facet target behavior prediction:}
Our manufacturing line features two sections prone to defects: pre-assembly and assembly. Pre-assembly requires {CT1} seconds to complete with an Overall Equipment Effectiveness (OEE) rate of {OEE1}\%. Assembly takes {CT2} seconds and has an OEE rate of {OEE2}\%. To cut total assembly time by 4 seconds, we must decide which section's duration can be reduced with the least increase in defects. Reducing cycle times will also result in higher line headcount costs. We have three strategy levels for decision-making: \\ \\

\noindent Novice strategy (targets encoded as 0 for pre-assembly, 3 for assembly): Intuitive choice.\\ \\

\noindent Intermediate strategy (targets encoded as 1 for pre-assembly, 4 for assembly): Make decision using key metrics.\\ \\

\noindent Expert strategy (targets encoded as 2 for pre-assembly, 5 for assembly): make well-informed judgments based on a comprehensive understanding of all relevant metric.\\ \\

\noindent Question: Given the different strategy levels, which options would you choose?\\ \\

\noindent Answer: 
\end{llmprompt}

\subsection{LLM Conversation Examples}
\label{sec:app:llm_prompts}
Both GPT-4 and Google Gemini deliver contradictory recommendations across prompts, despite the chain-of-thought reasoning in each case being similar and seemingly plausible.\\

\begin{llmprompt}{gray}{\paragraph{Shared System Prompt} Prompt used for generating LLM responses}
Our manufacturing line has two sections with potential defect sources: pre-assembly and assembly. Pre-assembly takes 40 seconds with an Overall Equipment Effective (OEE) rate of 88\%, while assembly takes 44 seconds with an OEE rate of 80.1\%. To reduce total assembly time by 4 seconds, we need to identify which section can be shortened with minimal defect increase. It's important to note that reducing cycle time will also lead to an increase in headcount costs. There are two options: reduce pre-assembly time or reduce assembly time. Which section will you choose to reduce time?
\end{llmprompt}

\begin{llmresponse}{blue}{\paragraph{Google Gemini} Response 1}
Reducing \textbf{pre-assembly} time is likely the better option in this scenario. Here's why:
\begin{itemize}
    \item \textbf{\textit{Higher OEE:}} Pre-assembly has a higher Overall Equipment Effectiveness (OEE) rate of 88\% compared to assembly's 80.1\%. This indicates that pre-assembly has a smaller inherent defect rate, and speeding it up is less likely to introduce additional defects.
    \item Smaller time reduction required: Reducing pre-assembly time by 4 seconds (40 seconds - 4 seconds = 36 seconds) achieves the overall target compared to needing a 4.4 second reduction in assembly time (44 seconds - 4 seconds = 39.6 seconds). A smaller time reduction generally translates to less disruption and less likelihood of introducing defects.
\end{itemize}
Therefore, focusing on optimizing the pre-assembly process to achieve a 4-second reduction has a better chance of success with minimal defect increase.
Here are some additional factors to consider:
\begin{itemize}
    \item Complexity of the processes: If pre-assembly is a more complex process with more steps, it might be harder to optimize without impacting quality.
    \item Automation potential: Analyze if either process can be automated further. Automation can improve consistency and potentially reduce cycle time.
\end{itemize}
It's always recommended to pilot any changes in a controlled environment before full-scale implementation. This allows you to measure the actual impact on cycle time, defect rates, and headcount requirements.
\end{llmresponse}

\begin{llmresponse}{blue}{\paragraph{Google Gemini} Response 2}
Analyzing the available information, shortening assembly time would likely be the better option to achieve a 4-second reduction in total \textbf{assembly} time while minimizing defect impact. Here's the reasoning:
\begin{itemize}
    \item \textbf{\textit{Higher OEE in Pre-Assembly:}}The OEE rate for pre-assembly (88\%) is higher than assembly (80.1\%). OEE reflects how efficiently a process utilizes available production time, factoring in quality and speed. A higher OEE suggests pre-assembly is already closer to optimal performance, and squeezing out 4 seconds of improvement might be more difficult without impacting quality.
    \item Potential for Improvement in Assembly: The lower OEE in assembly indicates more room for improvement without necessarily causing more defects. Optimizing this section might yield the needed 4-second reduction more readily.
\end{itemize}
Here's a breakdown of the thought process:
\begin{enumerate}
    \item Available Time for Improvement: OEE helps calculate the available time for improvement within a process. A higher OEE signifies less available time for improvement without sacrificing quality.
    \item Prioritizing Efficiency with Quality: Since minimizing defect rates is crucial, focusing on the section with a higher OEE (pre-assembly) for improvement might be riskier.
\end{enumerate}
Additional Considerations:
\begin{itemize}
    \item Analyze Defect Rates: While OEE is a good indicator, analyzing actual defect rates per section would provide a clearer picture. Even with a lower OEE, if assembly's defect rate is already low, improvement might be achievable.
    \item Cost-Effectiveness: Consider the cost of implementing changes in each section. Optimizing assembly might involve minor adjustments, while pre-assembly changes could require more effort.
\end{itemize}
It's recommended to conduct a more detailed analysis of defect rates and potential improvement methods for both sections before making a final decision. However, based on the OEE data provided, prioritizing assembly time reduction seems like a promising initial approach to achieve the 4-second target while minimizing defect risks. 
\end{llmresponse}

\begin{llmresponse}{orange}{\paragraph{OpenAI GPT 4.0} Response 1}
To decide which section to focus on for reducing time while considering minimal defect increase and cost implications, let's analyze the options:
\begin{enumerate}
    \item Pre-assembly Section:
    \begin{itemize}
        \item Current cycle time: 40 seconds
        \item OEE rate: 88\%
    \end{itemize}
    \item Assembly Section:
    \begin{itemize}
        \item Current cycle time: 44 seconds
        \item OEE rate: 80.1\%
    \end{itemize}
\end{enumerate}
You want to reduce total assembly time by 4 seconds. Here are the implications of reducing time in each section:
Option 1: Reduce Pre-assembly Time
\begin{itemize}
    \item Current Pre-assembly Time: 40 seconds
    \item Potential Reduced Time: Let's assume you can reduce this by 2 seconds to 38 seconds.
\end{itemize}
Implications:
\begin{itemize}
    \item Time Saved: 2 seconds per unit at pre-assembly.
    \item Total Assembly Time Saved: Since assembly time needs to be reduced by 4 seconds, reducing pre-assembly by 2 seconds means you still need to reduce assembly time by another 2 seconds.
\end{itemize}
Option 2: Reduce Assembly Time
\begin{itemize}
    \item Current Assembly Time: 44 seconds
    \item Required Reduction: 4 seconds to meet the goal.
\end{itemize}
Implications:
\begin{itemize}
    \item Time Saved: 4 seconds per unit at assembly.
\end{itemize}
Considerations:
\begin{itemize}
    \item \textbf{\textbf{Defect Rate Increase:}} Typically, when cycle times are reduced, there is a risk of increased defects due to rushed processes or errors.
    \item Cost Implications: Reducing cycle time often requires additional investment, such as hiring more workers to maintain output or investing in more efficient equipment.
\end{itemize}
\textbf{Decision:}
Given the goal is to reduce total assembly time by 4 seconds with minimal defect increase and considering cost implications:
\begin{itemize}
    \item \textbf{Reducing Assembly Time:} This option directly meets the required reduction goal (4 seconds) without needing additional reduction efforts in the pre-assembly section.
    \item \textbf{Defect Risk:} While both sections pose a risk of increased defects with faster cycle times, assembly might be more critical as it directly affects the final product quality and functionality.
\end{itemize}
\textbf{Conclusion:}
I would choose to focus on reducing the cycle time in the \textbf{Assembly} Section. This choice aligns with the goal of reducing assembly time by 4 seconds and may be more straightforward in achieving the required reduction compared to trying to split the reduction between pre-assembly and assembly. However, careful monitoring of defects and quality control measures should accompany any cycle time reduction to mitigate risks effectively.
\end{llmresponse}

\begin{llmresponse}{orange}{\paragraph{OpenAI GPT 4.0} Response 2}
To determine which section to reduce time in order to minimize defect increase, let's analyze the impact of reducing cycle time in each section based on the given OEE (Overall Equipment Effectiveness) rates and defect considerations.
Definitions and Given Data:
\begin{itemize}
    \item Pre-assembly:
    \begin{itemize}
        \item Cycle Time: 40 seconds
        \item OEE: 88\%
    \end{itemize}
    \item Assembly:
    \begin{itemize}
        \item Cycle Time: 44 seconds
        \item OEE: 80.1\%
    \end{itemize}
\end{itemize}
OEE Formula:
OEE is given by: 
\[
\text{OEE} = \text{Availability} \times \text{Performance} \times \text{Quality}
\]
The quality rate can be interpreted as the proportion of good products out of total products produced. Therefore, the defect rate (D) for each section is:
\[
D_{\text{Pre-assembly}} = 1 - \text{OEE}_{\text{Pre-assembly}} = 1 - 0.88 = 0.12
\]
\[
D_{\text{Assembly}} = 1 - \text{OEE}_{\text{Assembly}} = 1 - 0.801 = 0.199
\]
\textbf{Impact of Time Reduction:}
\begin{enumerate}
    \item Reducing Pre-assembly Time:
    \begin{itemize}
        \item Original Pre-assembly Time: 40 seconds
        \item New Pre-assembly Time: 40 - x seconds (where x is the reduction amount)
        \item Defect Rate Impact: The defect rate due to time reduction in pre-assembly is minimal because the OEE is relatively high. Also, reducing pre-assembly time might not significantly affect the defect rate if the process is already running efficiently.
    \end{itemize}
    \item Reducing Assembly Time:
    \begin{itemize}
        \item Original Assembly Time: 44 seconds
        \item New Assembly Time: 44 - y seconds (where y is the reduction amount)
        \item Defect Rate Impact: Assembly has a lower OEE, which translates to a higher defect rate. Reducing time in a section with a lower OEE often leads to an increase in defects because the process might be less stable and more sensitive to changes.
    \end{itemize}
\end{enumerate}
\textbf{Recommendation:}
Given that reducing time in the section with a higher defect rate (Assembly) will likely lead to a more significant increase in defects compared to reducing time in the section with a higher OEE (Pre-assembly), the preferable option is:
\textbf{Reduce Pre-assembly Time.}
\textbf{Reasoning:}
\begin{itemize}
    \item \textbf{\textbf{Lower Defect Rate Increase:}} Pre-assembly has a higher OEE and thus a lower defect rate. Reducing its time is less likely to significantly increase the defect rate compared to reducing time in the Assembly section.
    \item \textbf{Operational Impact:} While reducing time in either section will increase headcount costs, reducing Pre-assembly time is less likely to destabilize the process compared to Assembly, which has a lower OEE and higher defect rate.
\end{itemize}
In summary, to reduce total assembly time by 4 seconds with minimal defect increase, it is more advantageous to reduce \textbf{Pre-assembly} time rather than Assembly time. 
\end{llmresponse}

\end{document}